\documentclass{article}

\usepackage{graphics}
\usepackage{graphicx}
\usepackage{wrapfig}
\usepackage{hyperref}
\usepackage{amsfonts}
\usepackage{algorithm,algorithmicx}
\usepackage{algpseudocode}
\usepackage{microtype}      
\usepackage{pdfpages}
\usepackage{multicol}
\usepackage{amsmath}
\usepackage{amssymb}
\usepackage{float}
\usepackage{amsthm}
\usepackage[toc,page]{appendix}

\usepackage[accepted]{icml2021}

\icmltitlerunning{Maximum Mutation Reinforcement Learning for Scalable Control}

\begin{document}

\twocolumn[
\icmltitle{Maximum Mutation Reinforcement Learning for Scalable Control}



\icmlsetsymbol{equal}{*}

\begin{icmlauthorlist}
\icmlauthor{Karush Suri}{to,goo}
\icmlauthor{Xiao Qi Shi}{ed}
\icmlauthor{Konstantinos Plataniotis}{to}
\icmlauthor{Yuri Lawryshyn}{to,goo}

\end{icmlauthorlist}

\icmlaffiliation{to}{Department of Electrical \& Computer Engineering, University of Toronto, Canada.}
\icmlaffiliation{goo}{Center for Management of Technology \& Entrepreneurship (CMTE)}
\icmlaffiliation{ed}{RBC Capital Markets}

\icmlcorrespondingauthor{Karush Suri}{karush.suri@mail.utoronto.ca}

\icmlkeywords{ESAC, mutation, AMT, policy}

\vskip 0.3in
]



\printAffiliationsAndNotice{} 

\begin{abstract}
    Advances in Reinforcement Learning (RL) have demonstrated data efficiency and optimal control over large state spaces at the cost of scalable performance. Genetic methods, on the other hand, provide scalability but depict hyperparameter sensitivity towards evolutionary operations. However, a combination of the two methods has recently demonstrated success in scaling RL agents to high-dimensional action spaces. Parallel to recent developments, we present the Evolution-based Soft Actor-Critic (ESAC), a scalable RL algorithm. We abstract exploration from exploitation by combining Evolution Strategies (ES) with Soft Actor-Critic (SAC). Through this lens, we enable dominant skill transfer between offsprings by making use of soft winner selections and genetic crossovers in hindsight and simultaneously improve hyperparameter sensitivity in evolutions using the novel Automatic Mutation Tuning (AMT). AMT gradually replaces the entropy framework of SAC allowing the population to succeed at the task while \textit{acting as randomly as possible}, without making use of backpropagation updates. In a study of challenging locomotion tasks consisting of high-dimensional action spaces and sparse rewards, ESAC demonstrates improved performance and sample efficiency in comparison to the Maximum Entropy framework. Additionally, ESAC presents efficacious use of hardware resources and algorithm overhead. A complete implementation of ESAC can be found at \href{https://karush17.github.io/esac-web/}{karush17.github.io/esac-web/}.
\end{abstract}

\section{Introduction}
    Concepts and applications of Reinforcement Learning (RL) have seen a tremendous growth over the past decade \citep{atari}. These consist of applications in arcade games \citep{atari}, board games \citep{go} and lately, robot control tasks \citep{ddpg}. A primary reason for this growth is the usage of computationally efficient function approximators such as neural networks \citep{imagenet}. Modern-day RL algorithms make use of parallelization to reduce training times \citep{a3c} and boost agent's performance through effective exploration giving rise to scalable methods \citep{SRL, dreamer, aktr}. However, a number of open problems such as approximation bias, lack of scalability in the case of long time horizons and lack of diverse exploration restrict the application of scalability to complex control tasks.  
    
    State-of-the-art RL algorithms such as Soft Actor-Critic (SAC) \citep{sac} maximize entropy which is indicative of continued exploration. However, using a computationally expensive framework limits scalability as it increases the number of gradient-based \citep{backpropagation} updates of the overall algorithm. Moreover, tasks consisting of long time horizons have higher computational overhead as a result of long trajectory lengths. For instance, obtaining accurate position estimates \cite{SRL} over longer horizons require additional computation times which varies \textit{linearly} with the hardware requirement. Such a variation calls for increased scalability in the RL domain. 
    
    Diverse exploration strategies are essential for the agent to navigate its way in the environment and comprehend intricate aspects of less visited states\cite{mmktd}. Various modern-day RL methods lack significant exploration \citep{a3c,atari} which is addressed by making use of meta-controller\citep{hdqn} and curiosity-driven \citep{rnd} strategies at the cost of sample efficiency and scalability.  
    	
    Recent advances in RL have leveraged evolutionary computing for effective exploration and scalability \citep{es,epg,erl,neat,backpropamine}. These methods often fall short of optimal performance and depict sensitivity towards their hyperparameters. A common alternative for improving performance is to combine gradient-based objectives with evolutionary methods \citep{erl}. These algorithms allow a population of learners to gain dominant skills \cite{merl} from modern-day RL methods and depict robust control while demonstrating scalability. However, their applications do not extend to high-dimensional tasks as a result of sensitivity to mutational hyperparameters which still remains an open problem. 
    
    We introduce the Evolution-based Soft Actor Critic (ESAC), an algorithm combining ES with SAC for state-of-the-performance equivalent to SAC and scalability comparable to ES. Our contributions are threefold;
\begin{itemize}
    \item ESAC abstracts exploration from exploitation by exploring policies in \textit
    {weight space} using evolutions and exploiting gradient-based knowledge using the SAC framework.
    \item ESAC makes use of soft winner selection function which, unlike prior selection criteria \citep{erl}, does not shield winners from mutation. ESAC carries out genetic crossovers in hindsight resulting in dominant skill transfer between members of the population. 
    \item ESAC introduces the novel Automatic Mutation Tuning (AMT) which maximizes the mutation rate of ES in a small clipped region and provides significant hyperparameter robustness without making use of backpropagation updates. 
\end{itemize}
    
    \section{Related Work}

    \subsection{Scalable Reinforcement Learning}
    Recent advances in RL have been successful in tackling sample-efficiency \cite{sac} and approximation bias (also known as overestimation bias) which stems from value of estimates approximated by the function approximator. Overestimation bias is a common phenomenon occurring in value-based methods \citep{doubleq,doubledqn,maxmin} and can be addressed by making use of multiple critics in \citep{td3} in the actor-critic framework \cite{a3c}. This in turn limits scalability of algorithms \cite{SRL} by increasing the number of gradient-based updates. Moreover, memory complexity of efficient RL methods increases linearly with the expressive power of approximators \cite{mem}, which in turn hinders scalability of RL to complex control tasks. 
    
    \subsection{Evolutionary Reinforcement Learning}
    Intersection of RL and Evolutionary methods has for long been studied in literature \citep{epg,neat,backpropamine,symbiotic,contopt,cem}. \citep{es} presents the large-scale parallelizable nature of Evolution Strategies (ES). Performance of ES on continuous robot control tasks in comparison to various gradient-based frameworks such as Trust Region Policy Optimization (TRPO) \citep{trpo} and Proximal Policy Optimization (PPO) \citep{ppo} have been found comparable. On the other hand, ES falls short of competitive performance resulting in local convergence and is extremely sensitive to mutation hyperparameters. 
    
    An alternative to a pure evolution-based approach is a suitable combination of an evolutionary algorithm with a gradient-based method \citep{cma}, commonly referred to as Evolutionary Reinforcement Learning (ERL) \citep{erl}. ERL makes use of selective mutations and genetic crossovers which allow weak learners of the population to inherit skills from strong learners while exploring. ERL methods are scalable to high-dimensional control problems including multi-agent settings \cite{merl}. Such an approach is a suitable trade-off between sample efficiency and scalability but does not necessarily introduce mutation robustness. Other methods in literature \citep{epg} follow a similar approach but are often limited to directional control tasks which require little mutation. Thus, addressing scalability and exploration while preserving higher returns and mutation robustness requires attention from a critical standpoint. Our work is parallel to prior efforts made towards this direction. 
    

\section{Background}

    \subsection{Reinforcement Learning and Soft Actor-Critic}
    We review the RL setup wherein an agent interacts with the environment in order to transition to new states and observe rewards by following a sequence of actions. The problem is modeled as a finite-horizon Markov Decision Process(MDP) \citep{rl} defined by the tuple $(\mathcal{S},\mathcal{A},r,P,\gamma)$ where the state space is denoted by $\mathcal{S}$ and action space by $\mathcal{A}$, $r$ presents the reward observed by agent such that $r: \mathcal{S} \times \mathcal{A} \rightarrow [r_{min},r_{max}]$, $P: \mathcal{S} \times \mathcal{S} \times \mathcal{A} \rightarrow [0,\infty)$ presents the unknown transition model consisting of the transition probability to the next state $s_{t+1} \in \mathcal{S}$ given the current state $s_{t} \in \mathcal{S}$ and action $a_{t} \in \mathcal{A}$ at time step $t$ and $\gamma$ is the discount factor. We consider a policy $\pi_{\theta}({a_{t}|s_{t}})$ as a function of model parameters $\theta$. Standard RL defines the agent's objective to maximize the expected discounted reward $\mathbb{E}_{\pi_{\theta}}[\sum_{t=0}^{T}\gamma^{t}r(s_{t},a_{t})]$ as a function of the parameters $\theta$. 
    SAC \citep{sac} defines an entropy-based\citep{entropy} objective expressed as in \autoref{eq:sac}. 
    \begin{gather}
    J(\pi_{\theta}) = \sum_{t=0}^{T}\gamma^{t}[r(s_{t},a_{t}) + \lambda \mathcal{H}(\pi_{\theta}(\cdot|s_{t}))] \label{eq:sac}
    \end{gather}

    wherein $\lambda$ is the temperature coefficient and $\mathcal{H}(\pi_{\theta}(\cdot|s_{t}))$ is the entropy exhibited by the policy $\pi(\cdot|s_{t})$ in $s_{t}$. For a fixed policy, the soft Q-value function can be computed iteratively, starting from any function $Q:\mathcal{S} \times \mathcal{A}$ and repeatedly applying a modified Bellman backup operator $\mathcal{T}^{\pi}$ given by \autoref{eq:bell}
    \begin{gather}
    \mathcal{T}^{\pi}Q(s_{t},a_{t}) = r(s_{t},a_{t}) + \gamma\mathbb{E}_{s_{t+1} \sim P}[V(s_{t+1})] \label{eq:bell}
    \end{gather}

    where $V(s_{t})$ is the soft state value function expressed in \autoref{eq:state-val}. 
    \begin{gather}
    V(s_{t}) = \mathbb{E}_{a_{t} \sim \pi}[Q(s_{t},a_{t}) - \log (\pi (a_{t}|s_{t}))] \label{eq:state-val}
    \end{gather}

    We consider a parameterized state value function $V_{\psi}(s_{t})$, a soft Q-function $Q_{\phi}(s_{t},a_{t})$ and a policy $\pi_{\theta}(a_{t}|s_{t})$ which can be represented with nonlinear function approximators such as neural networks with $\psi, \phi$ and $\theta$ being the parameters of these networks.     
    
    \subsection{Evolution Strategies}
    We review the Evolution Strategies \citep{es} framework which is motivated by natural evolution. ES is a heuristic search procedure in which a population of offsprings is mutated using random perturbations. Upon mutation, the fitness objective corresponding to each member of the population is evaluated and offsprings with greater scores are recombined to form the population for the next generation. Let $n$ be the number of offsprings in the population. The parameter vectors of the model can then be represented as $\theta_{es,(i)}$ such that $i=1,2,..n$. A total of $n$ random perturbations $\epsilon_{(i)}$, $i=1,2,..n$ are sampled from a Gaussian distribution $\mathcal{N}(0,1)$ in order to mutate $\theta_{es,(i)}$ and evaluate the fitness objective $\mathbb{E}_{\epsilon_{(i)} \sim \mathcal{N}(0,1)}[O(\theta_{es,(i)} + \sigma \epsilon_{(i)})] = \frac{1}{\sigma}\mathbb{E}_{\epsilon_{(i)} \sim \mathcal{N}(0,1)}[O(\theta_{es,(i)} + \sigma \epsilon_{(i)})\epsilon_{(i)}]$. Here, $\sigma$ is the mutation rate which controls the extent of mutation. In the case of RL, the fitness objecive $O(\theta_{es,(i)})$ is the episodic reward observed by members of the population. 
    
    \section{A Motivating Example: The Discrete Cyclic MDP}
    
    \begin{figure}[H]
        \centering
        \includegraphics[width=4cm]{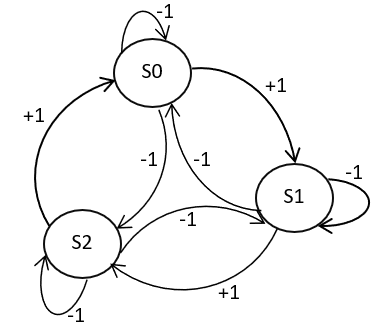}
        \caption{The long-horizon discrete Cyclic MDP. The agent observes a reward of +1 for moving clockwise and -1 otherwise. }
        \label{fig:mdp}
    \end{figure}
    
    We consider a long-horizon discrete cyclic MDP as our motivation for the work. The MDP has a state space $\mathcal{S}^{3}$ consisting of 3 states- $S0, S1$ and $S2$ and a discrete action space $\mathcal{A}^{3}$ consisting of 3 actions- clockwise, anticlockwise and stay. The agent starts in state $S0$. The reward function $r: \mathcal{S}^{3} \times \mathcal{A}^{3}$ assigns a reward of $+1$ for moving clockwise and $-1$ otherwise. Each episode lasts $2000$ timesteps and terminates if the agent reaches the end of horizon or incurs a negative reward. \autoref{fig:mdp} presents the long-horizon discrete cyclic MDP. 

\begin{figure*}[ht]
    \centering
    \includegraphics[width=13cm,height=4cm]{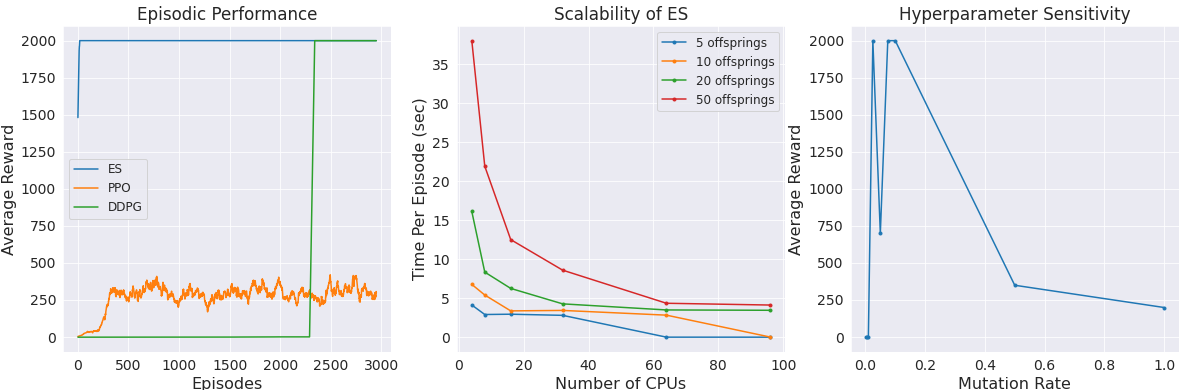}
    \caption{Comparison of ES, DDPG and PPO in the discrete cyclic MDP. ES depicts sample-efficient behavior (left) due to the presence of strong-learners in the population. Scalable nature of ES (center) with readily available computational resources allows in reduction of average episode execution time. However, ES is sensitive to hyperparameters (right) which results inconsistency across different seeds and rigorous fine-tuning.}
    \label{fig:mdp_results}
\end{figure*}

    The cyclic MDP, being a long-horizon problem, serves as a suitable benchmark for agent's behavior consisting of minimum computational overhead and is a small-scale replication of policy-search for scalable policy-based agents. The environment consists of a global objective which the agents can achieve if they solve the environment by obtaining the maximum reward of $+2000$. In order to assess evolution-based behavior, we compare the performance of a population of 50 offsprings utilizing ES with PPO \citep{ppo} and Deep Deterministic Policy Gradient (DDPG) \citep{ddpg}, an efficient off-policy RL method. Although DDPG is primarily a continuous control algorithm, we employ it as a result of the minimal nature of the problem. \autoref{fig:mdp_results} (left) presents the performance of ES in comparison to gradient-based agents in the cyclic MDP averaged over 3 runs. The ES population presents sample efficiency by solving the task within the first 100 episodes. DDPG, on the other hand, starts solving the task much latter during training. The use of a deterministic policy allows DDPG to continuously move left whereas in the case of ES, the population carries out exploration in the weight space and moves along the direction of the strong learners. Lastly, PPO finds a local solution and does not converge towards solving the task. Driven by clipped updates, PPO restricts the search horizon in policy space leading to a sub-optimal policy. 
    
    ES has proven to be scalable to large-scale and high dimensional control tasks \citep{es}. We assess this property of ES in the cyclic MDP by varying the operational hardware (number of CPUs) \citep{seed} and algorithm overhead (population size). We measure the average wall-clock time per episode \cite{nas}. As shown in \autoref{fig:mdp_results} (center), ES is parallelizable in nature and can be scaled up to larger population sizes by reducing the computation time. The large-scale readily parallelizable nature of ES is a convincing characteristic for utilizing CPU-based hardware. However, ES relies on excessively sensitive hyperparameters such as mutation rate. \autoref{fig:mdp_results} (right) presents the sensitivity of ES to mutation rate within a small range with a constant population size of 50. Varying population size does not present a trend in sensitivity indicating that mutation rate is the dominant hyperparameter governing policy behavior among offsprings. Hyperparameter sensitivity requires attention in the case of RL applications such as for real-world continuous control \citep{robot}. These include excessive tuning of parameters and detailed ablation studies. The cyclic MDP highlights this sensitive nature of ES and serves as a motivating example for tackling sensitivity while preserving optimal performance in a scalable manner. 

\section{Evolution-based Continuous Control}

The motivation behind ESAC stems from translating the scalability and tackling the mutation sensitivity of ES observed in discrete cyclic MDP to continuous control tasks. ESAC combines the scalable nature of ES with the limited approximation bias of SAC to yield a CPU-friendly state-of-the-art equivalent algorithm. 

\subsection{Evolution-based Soft Actor-Critic}
    \begin{figure}[H]
        \centering
        \includegraphics[width=6cm]{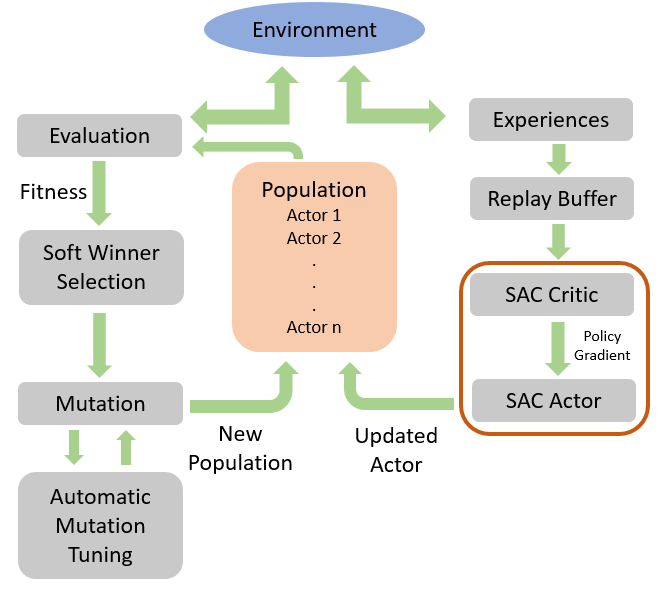}
        \caption{Workflow of ESAC combining ES with SAC. ESAC makes use of soft winner selections, hindsight crossovers and AMT for scalable performance.}
        \label{fig:schematic}
    \end{figure}
\subsubsection{Overview}
\autoref{fig:schematic} provides a high-level schematic of the ESAC algorithm and its components. The population is evaluated in the environment with the fitness metric as episodic rewards obtained by each offspring. Top $w$ winners are then segregated for mutation consisting of ES update followed by crossovers between perturbed offsprings and winners. The new population is formed using crossed-over offsprings and SAC agent. The SAC agent executes its own episodes at fixed timesteps and stores these experiences in a dedicated replay-buffer following policy update. During the SAC update timesteps, ESAC utilizes AMT which maximizes the mutation rate in a clipped region. SAC update timesteps are exponentially annealed to reduce entropy noise and abstract exploration in weight space.
    
\subsubsection{Algorithm}    
Algorithm \ref{alg:algorithm1} presents the ESAC algorithm. We begin by initializing $\psi$, $\theta$, $\theta_{es}$, $\phi$ being the parameters of state-value function, SAC policy, ES policy and Q-function respectively. We then initialize learning rate for SAC agent $\alpha$, learning rate of ES population $\alpha_{es}$, mutation rate $\sigma$, $p_{sac}$ which is the probability of SAC updates, $\bar{\psi}$ is the parameter vector of the target value function, $\zeta$ is the clip parameter, $\tau$ is the target smoothing coefficient, $e$ is the fraction of winners and $g$ is the gradient interval. A population of $n$ actors $pop_{n}$ is initialized along with an empty replay buffer $R$. Following the main loop, for each offspring $i$ in the population, we draw a noise vector $\epsilon_{i}$ from $\mathcal{N}(0,1)$ and perturb the ES policy vector $\theta_{es}$ to yield the perturbed parameter vector $\theta_{es,(i)}$ as per the expression $\theta_{es,(i)} + \sigma \epsilon_{(i)}$. $\theta_{es,(i)}$ is then evaluated to yield the fitness $F_{(i)}$ as episodic rewards. These are collected in a normalized and ranked set $F$. We now execute soft winner selection wherein the first $w= (n*e)$ offsprings from $F$ are selected for crossovers by forming the set $W$. The soft winner selection allows dominant skill transfer between winners and next generation offsprings. Mutation is carried out using the ES update \citep{es}. SAC gradient updates are executed at selective gradient intervals $g$ during the training process. $p_{sac}$ is exponentially annealed to reduce entropy noise and direct exploration in the weight space. During each $g$, the agent executes its own episodes by sampling $a_{t} \sim \pi_{\theta}(a_{t}|s_{t})$, observing $r(s_{t}|a_{t})$ and $s_{t+1}$ and storing these experiences in $R$ as a tuple $(s_{t},a_{t},r(s_{t},a_{t}),s_{t+1})$. Following the collection of experiences, we update the parameter vectors $\psi$, $\phi$ and $\theta$ by computing $\nabla_{\psi}J_{V}(\psi)$,  $\nabla_{\phi}J_{Q}(\phi_{(i)})$ and $\nabla_{\theta}J_{\pi}(\theta)$ where $J_{V}(\psi)$, $J_{Q}(\phi_{(i)})$ and $J_{\pi}(\theta)$ are the objectives of the state-value function, each of the two Q-functions $i \in \{ 1,2\}$ and policy as presented in \cite{sac} respectively. Gradient updates are followed by AMT update(\autoref{AMT}) which leads to hindsight crossovers between winners in $W$ and ES policy parameter vector $\theta_{es}$. Crossovers are carried out as random replacements between elements of weight vectors. In the case of hindsight crossovers, replacements between weight vector elements of current \& immediate previous generations is carried out. This allows the generation to preserve traits of dominant offsprings in hindsight. Finally, the new population is formed using $\theta,\theta_{es}$ and $W$.
\begin{algorithm}[!h]
\caption{Evolution-based Soft Actor-Critic (ESAC)}
\label{alg:algorithm1}
\begin{algorithmic}[1]

  \State Initialize parameter vectors $\psi,\bar{\psi},\theta, \theta_{es},\phi$
  \State Initialize $\alpha$, $\alpha_{es}$, $\sigma$, $\zeta$, $\tau$, $e$, $g$, $p_{sac}$
  \State Initialize a population of $n$ actors $pop_{n}$ and an empty replay buffer $R$

  \For{generation=1,$\infty$}
    \For{$i \in pop_{n}$}
        \State sample $\epsilon_{(i)} \sim \mathcal{N}(0,1)$
        \State $F_{(i)} \leftarrow$ evaluate $(\theta_{es,(i)} + \sigma \epsilon_{(i)})$ in the environment
    \EndFor
    \State normalize and rank $F_{(i)} \in F$
    \State select the first $w = (n*e)$ offsprings from $F$ to form the set of winners $W$
    \State set $\theta_{es} \leftarrow \theta_{es} + \frac{\alpha_{es}}{n\sigma} \sum_{i=1}^{n} F_{(i)}\epsilon_{(i)}$


    \If{$generation$ mod $g$ == 0 \& $\mu \sim \mathcal{N}(0,1) < p_{sac}$}
        \For{each environment step} 
            \State $a_{t} \sim \pi_{\theta}(a_{t}|s_{t})$
            \State observe $r(s_{t}|a_{t})$ and $s_{t+1} \sim P$
            \State $R \leftarrow R \cup {(s_{t},a_{t},r(s_{t},a_{t}),s_{t+1})}$
        \EndFor
        \For{each gradient step}
            \State $\psi \leftarrow \psi  - \alpha \nabla_{\psi}J_{V}(\psi)$
            \State $\phi \leftarrow \phi  - \alpha \nabla_{\phi}J_{Q}(\phi_{(i)})$ for $i \in \{1,2\}$
            \State $\theta \leftarrow \theta  - \alpha \nabla_{\theta}J_{\pi}(\theta)$
            \State $\bar{\psi} \leftarrow \tau \psi + (1 - \tau)\psi$
        \EndFor
        \State Update $\sigma$ using \autoref{eq:3}
    \EndIf
    
    \State crossover between $\theta_{es,(i)}$ and $\theta_{es}$ for $i=1,2,..w$
    \State Form new population $pop_{n}$ using $\theta, \theta_{es},W$
  \EndFor  

\end{algorithmic}
\end{algorithm}

\subsection{Automatic Mutation Tuning (AMT)}
\label{AMT}
Maximization of randomness in the policy space is akin to maximization in the weight space as both formulations are a multi-step replica of generalized policy improvement algorithm. This allows one to leverage the more suitable weight space for parallel computations. Policy updates during execution of offsprings require tuning the exploration scheme. To this end, we automatically tune $\sigma$ with the intial value $\sigma_{(1)}$. $\sigma$ is updated at fixed timesteps in a gradient-ascent manner without making use of backpropagation updates. AMT motivates guided exploration towards the objective as a result of the expansion of the search horizon of population which in turn enables the agent to maximize rewards \textit{as randomly as possible}. AMT makes use of the SmoothL1 (Huber) \citep{huber} loss function provided in \autoref{eq:1} and the update rule is mathematically expressed in \autoref{eq:2}.
\begin{gather}
    SmoothL1(x_{i},y_{i}) = 
    \begin{cases}
    0.5(x_{i} - y_{i})^{2}, \text{if}\ |x_{i} - y_{i}| < 1 \\
    |x_{i} - y_{i}| -0.5, \text{otherwise}
    \end{cases} \label{eq:1}
\end{gather}
\begin{gather}
    \sigma_{(t+1)} \xleftarrow[]{} \sigma_{(t)} + \frac{\alpha_{es}}{n\sigma_{(t)}}SmoothL1(R_{max},R_{avg})  \label{eq:2}
\end{gather}
Here, $R_{max}$ is the reward observed by winner offspring, $R_{avg}$ is the mean reward of the population with $\sigma_{(t)}$ and $\sigma_{(t+1)}$ the mutation rates at timesteps $t$ and $t+1$ respectively. While exploring in weight space, the SmoothL1 loss tends to take up large values. This is indicative of the fact that the deviation between the winner and other learners of the population is significantly high. In order to reduce excessive noise from weight perturbations $\epsilon_{i}$, we clip the update in a small region parameterized by the new clip parameter $\zeta$. Suitable values for $\zeta$ range between $10^{-6}$ to $10^{-2}$. The clipped update is mathematically expressed in \autoref{eq:3}. 
\begin{gather}
    \sigma_{(t+1)} \xleftarrow[]{} \sigma_{(t)} + clip(\frac{\alpha_{es}}{n\sigma_{(t)}}SmoothL1(R_{max},R_{avg}),0, \zeta)  \label{eq:3}
\end{gather}
The update can be expanded recursively and written in terms of the initial mutation rate $\sigma_{(1)}$ as expressed in \autoref{eq:4} (derived in \autoref{sc:derivation}).
\begin{gather}
    \theta_{(t+1)} \xleftarrow[]{} \theta_{(t)} + \frac{\alpha_{es}}{n\sigma_{(1)}\hat{\Lambda}} SmoothL1(R_{max,(t)},R_{avg,(t)}) \label{eq:4}
\end{gather}
Here, $\hat{\Lambda}$ is defined as the Tuning Multiplier and can be mathematically expressed as in \autoref{eq:tuning}. We direct the curious reader to \autoref{sc:derivation} for a full derivation.
\begin{multline}
    \hat{\Lambda} = \prod_{t^{'}=1}^{t-1} (1 + \frac{\alpha_{es}}{n \sigma_{(t^{'})}^{2}} SmoothL1(R_{max,(t^{'})},R_{avg,(t^{'})})) \label{eq:tuning}
\end{multline}

It can be additionally shown that AMT, when combined with soft winner selection, leads to policy improvement with high probability among the set of winners. We defer this proof to \autoref{sc:policy}.

\section{Experiments}

\begin{table*}[ht]
\resizebox{\textwidth}{!}{
\centering
 \begin{tabular}{c c c c c c c} 
 \hline
 Domain & Tasks & ESAC & SAC & TD3 & PPO & ES \\ [0.5ex] 
 \hline
 {} & HalfCheetah-v2 & 10277.16$\pm$403.63 & \textbf{10985.90$\pm$319.56} & 7887.32$\pm$532.60 & 1148.54$\pm$1455.64 & 3721.85$\pm$371.36\\ 
 {} & Humanoid-v2 & 5426.82$\pm$229.24 & \textbf{5888.55$\pm$44.66} & 5392.89$\pm$363.11 & 455.09$\pm$213.88 & 751.65$\pm$95.64\\
 {} & Ant-v2 & 3465.57$\pm$337.81 & 3693.08$\pm$708.56 & \textbf{3951.76$\pm$370.00} & 822.34$\pm$15.76 & 1197.69$\pm$132.01\\ 
 {} & Walker2d-v2 & \textbf{3862.82$\pm$49.80} & 3642.27$\pm$512.59 & 3714.89$\pm$90.35 & 402.33$\pm$27.38 & 1275.93$\pm$243.78\\
 MuJoCo & Swimmer-v2 & \textbf{345.44$\pm$17.89} & 31.68$\pm$0.41 & 110.85$\pm$23.02 & 116.96$\pm$0.74 & 254.42$\pm$109.91\\ 
 {} & Hopper-v2 & \textbf{3461.63$\pm$118.61} & 3048.69$\pm$467.21 & 3255.27$\pm$184.18 & 1296.17$\pm$1011.95 & 1205.73$\pm$185.25\\ 
 {} & LunarLanderContinuous-v2 & \textbf{285.79$\pm$9.60} & 66.52$\pm$26.75 & 273.75$\pm$4.51 & 124.47$\pm$11.58 & 74.41$\pm$109.69\\ 
 {} & Reacher-v2 & -2.01$\pm$0.07 & -0.50$\pm$0.05 & -5.12$\pm$0.17 & \textbf{-0.21$\pm$0.07} & -4.43$\pm$2.06\\ 
 {} & InvertedDoublePendulum-v2 & \textbf{9359.35$\pm$0.60} & 9257.96$\pm$86.54 & 5603.72$\pm$3213.51 & 88.52$\pm$4.73 & 259.39$\pm$36.75\\ 
 \hline
 {} & HumanoidStand & \textbf{805.08$\pm$135.67} & 759.08$\pm$125.67 & 745.15$\pm$291.377 & 8.41$\pm$3.33 & 10.57$\pm$0.30\\ 
 {} & HumanoidWalk & \textbf{883.00$\pm$21.97} & 843.00$\pm$7.97 & 686.33$\pm$56.23 & 2.20$\pm$0.18 & 10.59$\pm$0.34\\ 
 {} & HumanoidRun & \textbf{358.82$\pm$101.12} & 341.45$\pm$18.14 & 291.82$\pm$2101.12 & 2.29$\pm$0.16 & 10.55$\pm$0.30\\ 
 {DeepMind Control Suite} & CheetahRun & \textbf{773.14$\pm$3.00} & 227.66$\pm$13.07 & 765.22$\pm$27.93 & 371.70$\pm$19.82 & 368.62$\pm$32.87\\ 
 {} & WalkerWalk & \textbf{971.02$\pm$2.87} & 175.75$\pm$15.51 & 941.45$\pm$27.01 & 316.54$\pm$79.54 & 308.94$\pm$44.36\\ 
 {} & FishUpright & 914.96$\pm$2.04 & 285.69$\pm$21.03 & 838.32$\pm$34.86 & 561.39$\pm$111.59 & \textbf{997.58$\pm$0.26}\\ 
 \hline
 \end{tabular}}
 \caption{Average returns on 15 locomotion tasks from MuJoCo \& DeepMind Control Suite. Results are averaged over 5 random seeds with the best performance highlighted in bold. ESAC demonstrates improved performance on 10 out of 15 tasks. Furthermore, ESAC presents consistency across different seeds in the case of large action spaces and sparse rewards indicating the suitability of evolutionary methods to RL and control tasks.}
\label{tab:table}
\end{table*}

\begin{figure*}[ht]
    \centering
    \includegraphics[width=2cm,height=2cm]{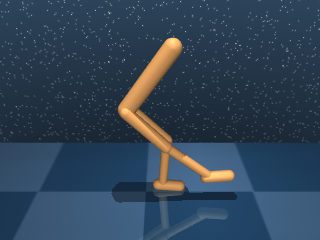}    \hspace{0.1cm}
    \includegraphics[width=2cm,height=2cm]{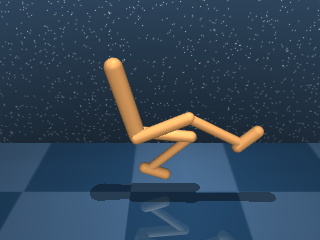}
    \hspace{0.1cm}
    \includegraphics[width=2cm,height=2cm]{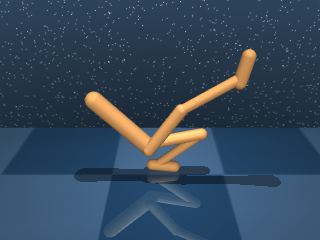}
    \hspace{0.1cm}
    \includegraphics[width=2cm,height=2cm]{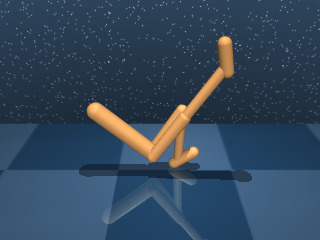}
    \hspace{0.1cm}
    \includegraphics[width=2cm,height=2cm]{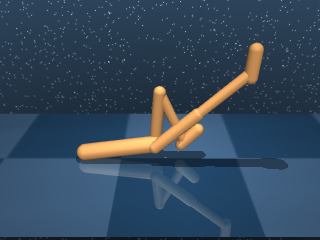}
    \hspace{0.1cm}
    \includegraphics[width=2cm,height=2cm]{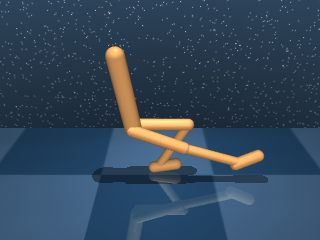}
    \hspace{0.1cm}
    \includegraphics[width=2cm,height=2cm]{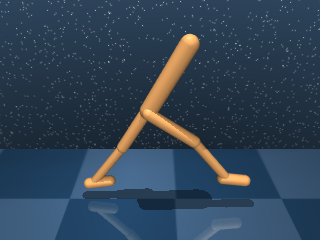}
    \caption{Robust behavior of ESAC observed on the WalkerWalk task. The ESAC policy prevents the walking robot from falling down when the robot loses its balance while walking. The robot successfully retains its initial posture within 10 timesteps. ESAC exhibits robust policies on complex tasks as a result of successive evolutions and hindsight genetic crossovers between winners and actors of the population.}
    \label{fig:robust}
\end{figure*}

Our experiments aim to evaluate performance, sample efficiency, scalability and mutation sensitivity of ESAC. Specifically, we aim to answer the following questions- 
\begin{itemize}
\item How does the algorithm compare to modern-day RL methods for complex tasks?
\item How do evolutionary operations impact scalability in the presence of gradients?
\item Which components of the method contribute to sensitivity and scalability? 
\end{itemize}

\subsection{Performance}
We assess performance and sample efficiency of ESAC with state-of-the-art RL techniques including on-policy and off-policy algorithms. We compare our method to ES \citep{es}; SAC \citep{sac}; Twin-Delayed Deep Deterministic Policy Gradient (TD3)\citep{td3} and PPO \citep{ppo} on a total of 9 MuJoCo \citep{mujoco} and 6 DeepMind Control Suite \citep{dm} tasks. We refer the reader to \autoref{sc:dm_results} for complete results. The tasks considered consist of sparse rewards and high-dimensional action spaces including 4 different versions of Humanoid. Additionally, we consider the LunarLander continuous task as a result of its narrow basin of learning. All methods were implemented using author-provided implementations except for ES in which Virtual Batch Normalization \cite{vbnorm} was omitted as it did not provide significant performance boosts and hindered scalability. 

Agents were trained in OpenAI's Gym environments \citep{gym} framework for a total of 5 random seeds. Training steps were interleaved with validation over 10 episodes. For all agents we use nonlinear function approximators as neural networks in the form of a multilayer architecture consisting of 2 hidden layers of 512 hidden units each activated with ReLU \citep{relu} nonlinearity and an output layer with tanh activation. We use this architecture as a result of its consistency in baseline implementations. We use Adam \citep{adam} as the optimizer (refer to \autoref{sc:hyperparameters} for hyperparameters). For ESAC and SAC, we use a Diagonal Gaussian (DG) policy \cite{curl} without automatic entropy tuning for a fair comparison. Training of gradient-based methods was conducted on 4 NVIDIA RTX2070 GPUs whereas for ES and ESAC, a total of 64 AMD Ryzen 2990WX CPUs were used. 

\autoref{tab:table} presents total average returns of agents on all 15 tasks considered for our experiments. ESAC demonstrates improved returns on 10 out of 15 tasks. ESAC makes use of evolution-based weight-space exploration to converge to robust policies in tasks where SAC often learns a sub-optimal policy. Moreover, utilization of evolutionary operations demonstrates consistency across different seeds for high-dimensional Humanoid tasks indicating large-scale suitability of the method to complex control.

\subsection{Behaviors}
Combination of RL and evolutionary methods provides suitable performance on control benchmarks. It is essential to assess the behaviors learned by agents as a result of weight-space exploration. We turn our attention to observe meaningful patterns in agent's behavior during its execution in the environment. More specfically, we aim to evaluate the robustness of ESAC scheme which promises efficacious policy as a result of effective exploration. We initialize a learned ESAC policy on the WalkerWalk task and place it in a challenging starting position. The Walker agent stands at an angle and must prevent a fall in order to complete the task of walking suitably as per the learned policy. \autoref{fig:robust} demonstrates the behavior of the Walker agent during its first 100 steps of initialization. The agent, on its brink of experiencing a fall, is able to gain back its balance and retain the correct posture for completing the walking task. More importantly, the agent carries out this manoeuvre within 50 timesteps and quickly gets back on its feet to start walking. \autoref{fig:robust} is an apt demonstration of robust policies learned by the ESAC agent. Dominant skill transfer arising from hindsight crossovers between winners and offsprings provisions effective exploration in weight space.  

\subsection{Scalability}

\begin{figure*}[ht]
    \centering
    \includegraphics[height=10cm,width=13.5cm]{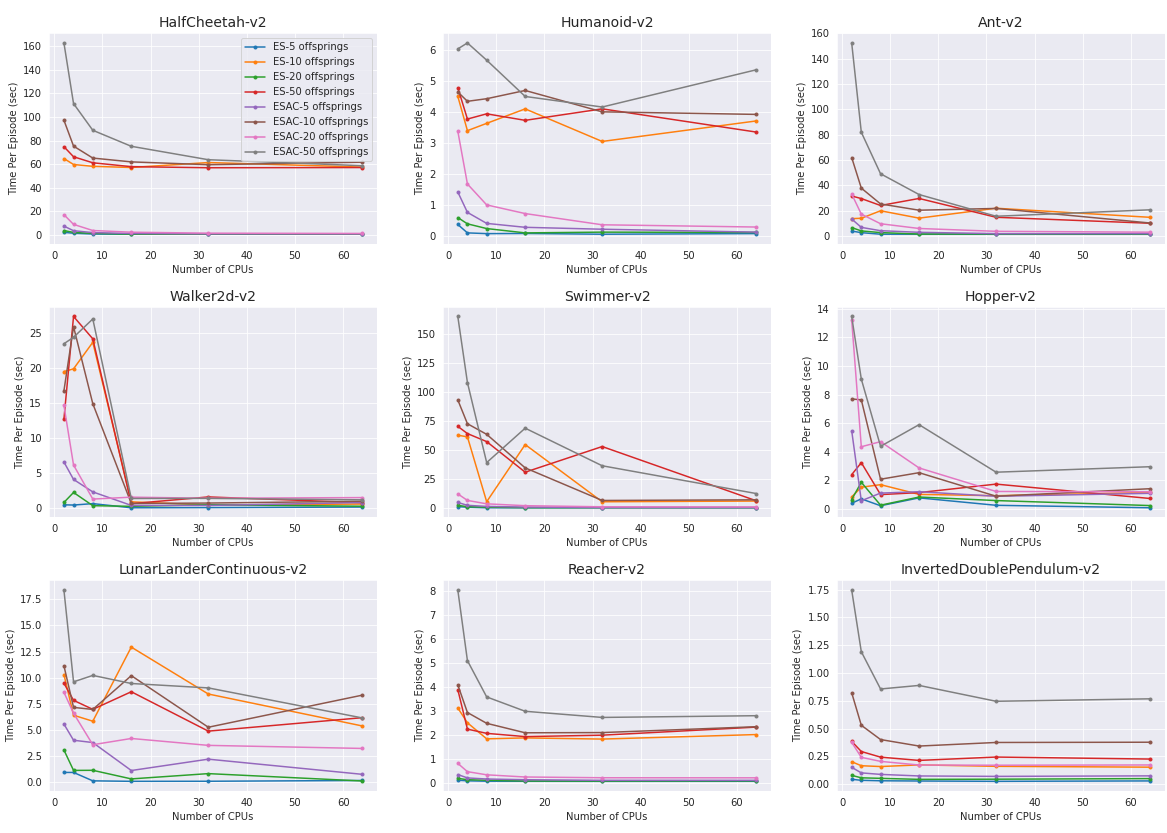}
    \caption{Variation of average time per episode (in seconds) with the number of operational CPUs and population size (in legend) for locomotion tasks from the MuJoCo benchmark. ESAC demonstrates equivalent scalability as ES by providing an approximate $60\%$ reduction in episode execution time. Reduction in computational time is found to be approximately quadratic which is computationally efficient for RL in comparison to linear variations for high-end GPU machines.}
    \label{fig:all_scale}
\end{figure*}

We assess scalability of our method with ES on the basis of hardware resources and algorithm overhead. We vary the number of CPUs by keeping other training parameters constant. Parallelization on multiple CPU-based resources is readily available and cost-efficient in comparison to a single efficient GPU resource. We also vary number of offsprings in the population by fixing CPU resources. Out of mutation rate $\sigma$ and population size $n$, $n$ governs the computational complexity of ES with $\sigma$ being a scalar value. Thus, assessing variation w.r.t $n$ provides a better understanding of resource utility and duration. For both experiments, we train the population for $10^{6}$ steps and average the wall-clock time per episode. Another effective way to demonstrate scalability is by monitoring overall time taken to complete the training tasks \citep{es}. However, this often tends to vary as initial learning periods have smaller episode lengths which does not compensate for fixed horizons of $1000$ steps in MuJoCo. 

\autoref{fig:all_scale} presents the scalable nature of ESAC equivalent to ES on the MuJoCo and LunarLanderContinuous tasks. Average wall-clock time per episode is reduced utilizing CPU resources which is found to be favourable for evolution-based methods. Moreover, the variation depicts consistency with the increasing number of members in the population indicating large-scale utility of the proposed method. A notable finding here is that although ESAC incorporates gradient-based backpropagation updates, it is able to preserve its scalable nature by making use of evolutions as dominant operations during the learning process. This is in direct contradiction to prior methods \cite{erl}which demonstrate reduced sample-efficiency and the need for significant tuning when combining RL with scalable evolutionary methods. Reduction in the number of SAC updates by exponentially annealing the gradient interval allows ESAC to reduce computation times and simultaneously explore using AMT.

\subsection{Ablation Study}

\begin{figure}[H]
    \centering
    \includegraphics[height=2.5cm,width=3.25cm]{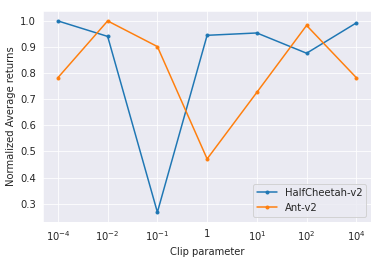}
    \label{fig:clip}
    \includegraphics[height=2.75cm,width=4cm]{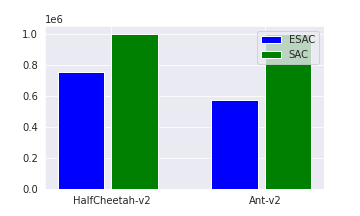}
    \caption{Left: Mutation sensitivity of ESAC, Right: Number of backprop updates in ESAC compared to SAC}
    \label{fig:updates}
\end{figure}

\subsubsection{Mutation Sensitivity}
 ES presents sensitivity to $\sigma$ which is addressed by making use of AMT in ESAC. The AMT update gradually increases mutation rate $\sigma$ using clip parameter $\zeta$ as learning progresses. \autoref{fig:clip} presents the variation of average normalized rewards with different values of the new hyperparameter $\zeta$ for HalfCheetah-v2 and Ant-v2 tasks. Each experiment was run for 1 million steps. The population presents robustness and performance improvement for small values with the optimal range being $10^{-4}$ to $10^{-2}$. On the other hand, sensitivity is observed in the $10^{-1}$ to $1$ region which accounts for larger updates with high variance. Hyperparameter variation is limited to a smaller region, in contrast to a wider spread of $\sigma$ in the ES update. Offsprings remain robust to significantly large values of $\zeta$ due to early convergence of the population at the cost of poor performance among weak learners of the population. However, this is addressed by making use of hindsight crossovers which allow simultaneous transfer of dominant traits.

\subsubsection{Number of Updates} 
The main computation bottleneck in SAC arises from the number of backprop updates. This is tackled by exponentially annealing these updates and increasing winner-based evolutions and crossovers for transferring skills between SAC agent and ES offspings. \autoref{fig:updates} presents a comparison between the number of backprop updates carried out using SAC and ESAC during the training phase of HalfCheetah-v2 and Ant-v2 tasks. Results for the updates are averaged over 3 random seeds. ESAC executes lesser number of updates highlighting its computationally efficient nature and low dependency on a gradient-based scheme for monotonic improvement. Complete results can be found in \autoref{sc:scale_results}.

\section{Conclusions}
In this paper, we introduce ESAC which combines the scalable nature of ES with low approximation bias and state-of-the-art performance of SAC. ESAC addresses the problem of mutation-sensitive evolutions by introducing AMT which maximizes the mutation rate of evolutions in a small clipped region as the SAC updates are exponentially decayed. ESAC demonstrates improved performance on 10 out of 15 MuJoCo and DeepMind Control Suite tasks including different versions of Humanoid. Additionally, ESAC presents scalability comparable to ES, reducing the average wall-clock time per episode by approximately $60\%$, hence depicting its suitability for large-scale RL tasks involving continuous control. On the other hand, high variance of the SAC agent under sparse rewards requiring consistent optimal behavior hurts the performance of ESAC. This can be addressed by combining the framework with a meta-controller or using a more sophisticated architecture such as a master-slave framework \citep{ms}. We leave this for future work.    




\bibliography{example_paper}
\bibliographystyle{icml2021}

\newpage
\appendix
\onecolumn    



\section{Derivation} \label{sc:derivation}
The AMT and ES update rules are given as\\
\begin{gather}
    \sigma_{(t)} \xleftarrow[]{} \sigma_{(t-1)} + \frac{\alpha_{es}}{n \sigma_{(t-1)}}SmoothL1(R_{max,(t-1)},R_{avg,(t-1)}) \nonumber \\
    \theta_{(t+1)} \xleftarrow[]{} \theta_{(t)} + \frac{\alpha_{es}}{n \sigma_{(t)}}\sum_{i=1}^{N}R_{i}\epsilon_{i} \nonumber
\end{gather}
Using the expression for $\sigma_{(t)}$ in the ES update yields the following\\
\begin{gather}
    \theta_{(t+1)} \xleftarrow[]{} \theta_{(t)} + \frac{\alpha_{es}}{n(\sigma_{(t-1)} + \frac{\alpha_{es}}{n \sigma_{(t-1)}}SmoothL1(R_{max,(t-1)},R_{avg,(t-1)}))}\sum_{i=1}^{n}R_{i}\epsilon_{i} \nonumber \\
    = \theta_{(t+1)} \xleftarrow[]{} \theta_{(t)} + \frac{\alpha_{es}}{n\sigma_{(t-1)}(1 + \frac{\alpha_{es}}{n \sigma_{(t-1)}^{2}}SmoothL1(R_{max,(t-1)},R_{avg,(t-1)}))}\sum_{i=1}^{n}R_{i}\epsilon_{i} \nonumber \\
    = \theta_{(t+1)} \xleftarrow[]{} \theta_{(t)} + \frac{\alpha_{es}}{n\sigma_{(t-1)}\Lambda_{(t-1)}} \sum_{i=1}^{n}R_{i}\epsilon_{i} \nonumber
\end{gather}
where $\Lambda_{(t-1)} = 1 + \frac{\alpha_{es}}{n \sigma_{(t-1)}^{2}}SmoothL1(R_{max,(t-1)},R_{avg,(t-1)})$. Expanding $\sigma_{(t-1)}$ using the AMT update rule gives us\\
\begin{gather}
    \theta_{(t+1)} \xleftarrow[]{} \theta_{(t)} + \frac{\alpha_{es}}{n\Lambda_{(t-1)}(\sigma_{(t-2)} + \frac{\alpha_{es}}{n \sigma_{(t-2)}}SmoothL1(R_{max,(t-2)},R_{avg,(t-2)}))}\sum_{i=1}^{n}R_{i}\epsilon_{i} \nonumber \\
    = \theta_{(t+1)} \xleftarrow[]{} \theta_{(t)} + \frac{\alpha_{es}}{n\Lambda_{(t-1)} \sigma_{(t-2)}(1 + \frac{\alpha_{es}}{n \sigma_{(t-2)}^{2}}SmoothL1(R_{max,(t-2)},R_{avg,(t-2)}))}\sum_{i=1}^{n}R_{i}\epsilon_{i} \nonumber \\
    = \theta_{(t+1)} \xleftarrow[]{} \theta_{(t)} + \frac{\alpha_{es}}{n\Lambda_{(t-1)}\sigma_{(t-2)}\Lambda_{(t-2)}} \sum_{i=1}^{n}R_{i}\epsilon_{i} \nonumber
\end{gather}
where $\Lambda_{(t-2)} = 1 + \frac{\alpha_{es}}{n \sigma_{(t-2)}^{2}}SmoothL1(R_{max,(t-2)},R_{avg,(t-2)})$. Expanding this recursively gives us the following\\
\begin{gather}
    \theta_{(t+1)} \xleftarrow[]{} \theta_{(t)} + \frac{\alpha_{es}}{n\sigma_{(1)} \Lambda_{(t-1)}\Lambda_{(t-2)} . . . \Lambda_{(1)}} \sum_{i=1}^{n}R_{i}\epsilon_{i} \nonumber \\
    = \theta_{(t+1)} \xleftarrow[]{} \theta_{(t)} + \frac{\alpha_{es}}{n\sigma_{(1)} \prod_{t^{'}=1}^{t-1} \Lambda_{(t^{'})}} \sum_{i=1}^{n}R_{i}\epsilon_{i} \nonumber \\
    = \theta_{(t+1)} \xleftarrow[]{} \theta_{(t)} + \frac{\alpha_{es}}{n\sigma_{(1)} \hat{\Lambda}} \sum_{i=1}^{n}R_{i}\epsilon_{i} \nonumber
\end{gather}

Hence, yielding the AMT update in the form of initial mutation rate $\sigma_{(1)}$. 

\section{Policy Improvement} \label{sc:policy}

We will now show that AMT improves the policy of winners in the population. Let us consider two successive gradient intervals $g$ indexed by $(l)$ and $(l-1)$. Let $p_{(l)}$ and $p_{(l-1)}$ be the probabilities of convergence to the optimal policy $\pi_{\theta_{es}}^{*}(a_{t}|s_{t})$ in the weight space at $(l)$ and $(l-1)$ respectively. \\

We start by evaluating the mutation rates at $(l)$ and $(l-1)$ which are given as $\sigma_{(l)} > \sigma_{(l-1)}$.
We can now evaluate the probabilities of convergence to $\pi_{\theta_{es}}^{*}(a_{t}|s_{t})$ as
\begin{gather}
    p_{(l)} \geq p_{(l-1)} \nonumber
\end{gather}
Using this fact, we can evaluate the winners (indexed by $q$) in the sorted reward population $F$.
\begin{gather}
    \sum_{q=1}^{w} p_{(l)}^{(q)} F_{(l)}^{(q)} \geq \sum_{q=1}^{w} p_{(l-1)}^{(q)} F_{(l-1)}^{(q)} \nonumber \\
    = \mathbf{E}_{F_{(l)}^{(q)} \sim F_{(l)}}[F_{(l)}^{(q)}] \geq \mathbf{E}_{F_{(l-1)}^{(q)} \sim F_{(l-1)}}[F_{(l-1)}^{(q)}] \nonumber \\
    = \mathbf{E}[W_{(l)}] \geq \mathbf{E}[W_{(l-1)}] \nonumber    
\end{gather}
Here, $p_{(l)}^{(q)}$ is the probability of convergence of actor $q$ (having observed reward $F_{(l)}^{(q)}$) to its optimal policy $\pi_{\theta_{es}}^{(q),*}(a_{t}|s_{t})$ at interval $(l)$. $W_{(l)}$ represents the set of winners at $(l)$. The mathematical expression obtained represents that the set of winners $W_{(l)}$ formed at the next gradient interval $(l)$ is at least as good as the previous set of winners $W_{(l-1)}$, i.e.- $\pi_{\theta_{es},(l)}^{(q)}(a_{t}|s_{t}) \geq \pi_{\theta_{es},(l-1)}^{(q)}(a_{t}|s_{t})$ . This guarantees policy improvement among winners of the population.

\section{Additional Results}
\subsection{Performance} \label{sc:dm_results}

\begin{figure}[H]
    \centering
    \includegraphics[width=17cm,height=9.5cm]{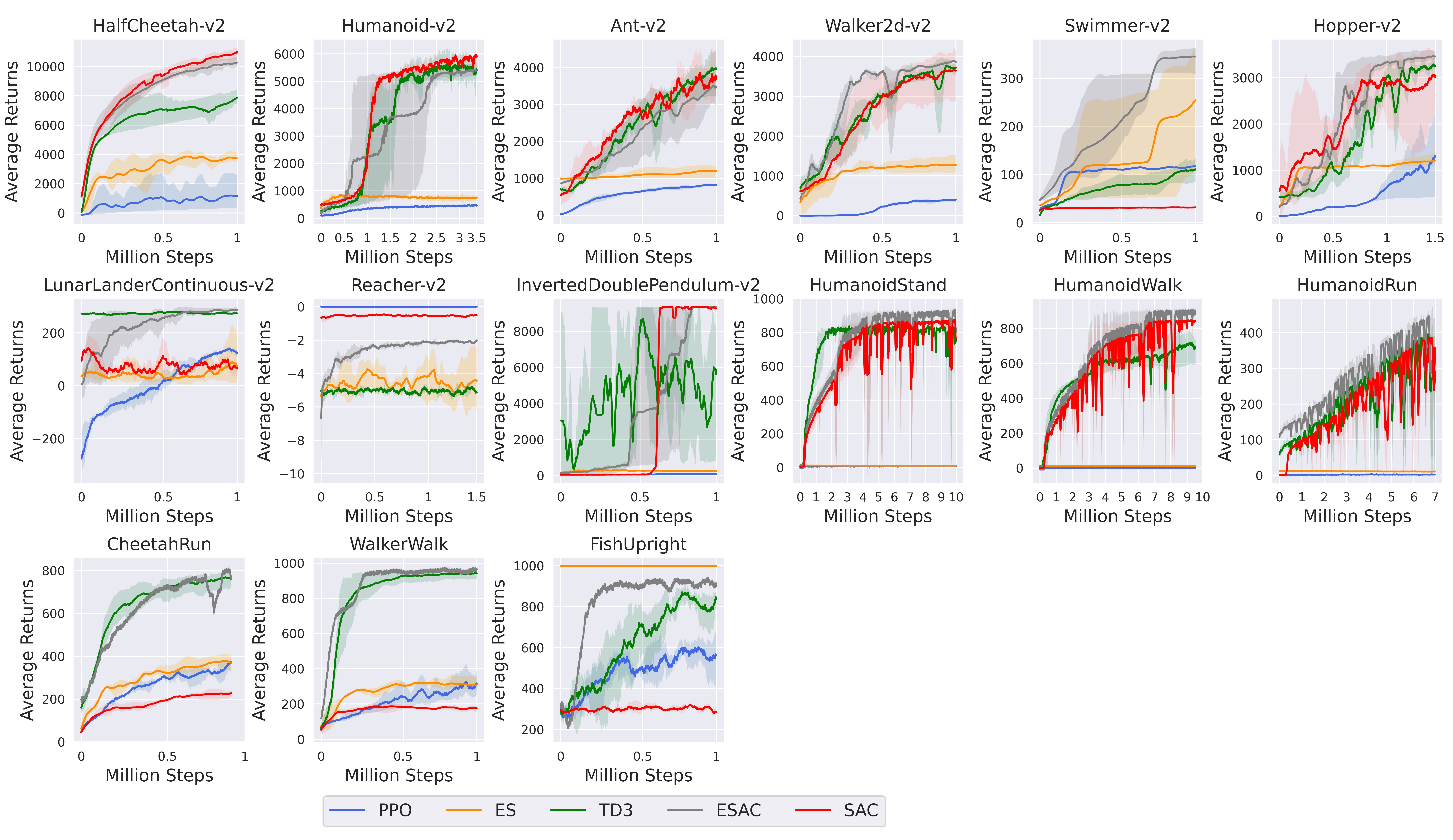}
    \caption{Average Returns on MuJoCo and DeepMind Control Suite tasks. ESAC's demonstrates improved performance on  out 10 of 15 tasks presented in \autoref{tab:table}.}
    \label{fig:rewards_dm}
\end{figure}

We evaluate the performance and sample-efficiency of ESAC on 9 MuJoCo and 6 DeepMind Control Suite \cite{dm} tasks. Figure \autoref{fig:rewards_dm} presents learning behavior of ESAC in comparison to SAC, TD3, PPO and ES on all 15 tasks. Training setup for all agents was kept same with different values of hyperparameters (presented in \autoref{sc:hyperparameters_dm}). ESAC demonstrates improved returns on 10 out of 15 tasks as presented in \autoref{tab:table}. Results are averaged over 5 random seeds with Humanoid experiments evaluated for 10 million steps.

\subsection{Scalability} \label{sc:scale_results}

\begin{figure}[H]
    \centering
    \includegraphics[height=5cm,width=14cm]{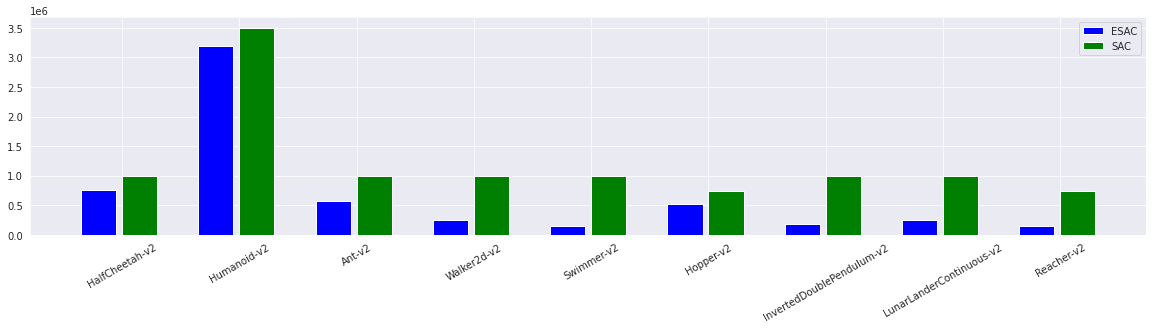}
    \caption{Complete results on the number of backprop updates for MuJoCo control tasks and LunarLanderContinuous environment from OpenAI's Gym suite. ESAC exponentially anneals gradient-based SAC updates and leverages winner selection and crossovers for computational efficiency and performance improvement.}
    \label{fig:scale}
\end{figure}

\section{Hyperparameters} \label{sc:hyperparameters}

\subsection{MuJoCo}

Hyperparameter values for our experiments are adjusted on the basis of complexity and reward functions of tasks. In the case of MuJoCo control tasks, training-based hyperparameters are kept mostly the same with the number of SAC episodes varying as per the complexity of task. Tuning was carried out on HalfCheetah-v2, Ant-v2, Hopper-v2 and Walker2d-v2 tasks. Out of these, Ant-v2 presented high variance indicating the requirement of a lower learning rate. Number of SAC updates in SAC and ESAC implementations were kept 1 for a fair comparison with TD3. All tasks have a common discount factor $\gamma=0.99$, SAC learning rate $\alpha= 3 \times 10^{-4}$, population size $n=50$, mutation rate $\sigma= 5 \times 10^{-3}$ and winner fraction $e=0.4$. ES learning rate $\alpha_{es}$ was kept fixed at $5 \times 10^{-3}$ for all tasks except Ant-v2 having $\alpha_{es} = 1 \times 10^{-4}$. \\

The only variable hyperparameter in our experiments is number of SAC episodes executed by the SAC agent. Although the ESAC population in general is robust to its hyperparameters, the SAC agent is sensitive to the number of episodes. During the tuning process, the number of episodes were kept constant to a value of 10 for all the tasks. However, this led to inconsistent results on some of the environments when compared to SAC baseline. As a result, tuning was carried out around this value to obtain optimal results corresponding to each task. The SAC agent executed a total of 10 episodes for each gradient interval $g$ for Ant-v2, Walker2d-v2, Hopper-v2 and Humanoid-v2 tasks; 5 episodes for HalfCheetah-v2, LunarLanderContinuous-v2, Reacher-v2 and InvertedPendulum-v2 tasks; and 1 episode for Swimmer-v2 task.

\subsection{DeepMind Control Suite} \label{sc:hyperparameters_dm}

The DeepMind control suite presents a range of tasks with sparse rewards and varying complexity for the same domain. Hyperparameter values for these tasks are different from that of MuJoCo control tasks. Tuning was carried on the Cheetah, Quadruped and Walker tasks in order to obtain sample-efficient convergence. In the case of SAC, granularity of hyperparameter search was refined in order to observe consistent behavior. However, different values of temperature parameter produced varying performance. As in the MuJoCo case, we kept the number of updates fixed to 1 in order to yield a fair comparison with TD3. \\

All tasks have a common discount factor $\gamma=0.99$, SAC learning rate $\alpha= 3 \times 10^{-4}$, population size $n=50$, mutation rate $\sigma= 1 \times 10^{-2}$, winner fraction $e=0.4$ and ES learning rate $1 \times 10^{-2}$. As in the case of MuJoCo tasks, the SAC is found to be sensitive to the number of episodes during the gradient interval $g$. These were kept constant at 5 and then tuned around this value for optimal performance. Final values of the SAC episodes were 5 for CartpoleSwingup, WalkerWalk and WalkerRun tasks and 1 for the remaining tasks.


\end{document}